\def\year{2021}\relax
\documentclass[letterpaper]{article} 
\usepackage{aaai21}  
\usepackage{times}  
\usepackage{helvet} 
\usepackage{courier}  
\usepackage[hyphens]{url}  
\usepackage{graphicx} 
\usepackage{natbib}  
\usepackage{caption} 
\usepackage{amsfonts}
\usepackage{eurosym}

\usepackage[switch]{lineno}  %
\title{Deep Multi-stream Network for Video-based Calving Sign Detection}
\author {
        Ryosuke Hyodo\textsuperscript{\rm 1}, 
        Teppei Nakano\textsuperscript{\rm 1,2},
        Tetsuji Ogawa\textsuperscript{\rm 1} \\
}
\affiliations {
    \textsuperscript{\rm 1}Waseda University, Tokyo, Japan\\
    \textsuperscript{\rm 2}Intelligent Framework Laboratory, Tokyo, Japan\\
    \{hyodo, teppei, ogawa\}@pcl.cs.waseda.ac.jp
}
\begin{document}
\maketitle

\begin{abstract}
We have designed a deep multi-stream network 
for automatically detecting calving signs from video.
Calving sign detection from a camera,
which is a non-contact sensor, 
is expected to enable more efficient livestock management.
As large-scale, well-developed data 
cannot generally be assumed
when establishing calving detection systems,
the basis for making the prediction needs to be presented to farmers during operation, so black-box modeling (also known as \textit{end-to-end} modeling) is not appropriate.
For practical operation of calving detection systems,
the present study aims to incorporate expert knowledge into a deep neural network.
To this end,
we propose a multi-stream calving sign detection network
in which multiple calving-related features are extracted
from the corresponding feature extraction networks 
designed for each attribute 
with different characteristics,
such as a cow's posture, rotation, and movement,
known as calving signs,
and are then integrated 
appropriately depending on the cow's situation.
Experimental comparisons conducted 
using videos of 15 cows demonstrated that
our multi-stream system yielded a significant improvement
over the \textit{end-to-end} system,
and the multi-stream architecture significantly contributed
to a reduction in detection errors.
In addition,
the distinctive mixture weights 
we observed helped provide interpretability of the system's behavior.
\end{abstract}

\section{Introduction}

In the management of breeding cattle,
assistance during calving is important
to prevent fatal accidents, such as stillbirth and dystocia.
The financial losses associated with dystocia are estimated to be \euro500 per case~\cite{McGuirk685}. 
This is also considered a welfare issue,
as dystocia is an extremely painful condition for cows~\cite{MEE2011189}.
Systems that notify farmers when calving is likely to start, 
therefore, are necessary for efficient livestock management.

In order to provide assistance at the early stage, 
contact-type sensors have been widely studied and commercialized~\cite{Ruttten, Marchesi, Sakatani, Krieger}.
For example, the Moocall Calving Sensor~\footnote{https://moocall.com} is an accelerometer that detects tail raising, 
which usually occurs before calving.
However, the installation of contact-type sensors,
which need to be attached directly on cows, 
seems burdensome to both cows and farmers.
In fact, 
80\% of interviewed farmers stated that
the cows’ behavioural reaction was negative
when such sensors were attached. 
Farmers also observed damage to the tails after
using the sensor,
and 20\% reported such severe damage as to require amputation~\cite{Lind1372909}.
Moreover, the information that a contact-type sensor can acquire is limited; \textit{e.g.,}
a sensor that recognizes the tail raising cannot capture an increase in step count, which is also a known calving sign.
A camera, in contrast, which is a non-contact sensor, 
can easily capture information on multiple attributes.
Thus, detecting calving signs using cameras
is desirable for both human- and cow-friendly surveillance, 
and is also scalable to handle multiple calving signs.

\begin{figure}[tb]
\centering
\includegraphics[width=\linewidth]{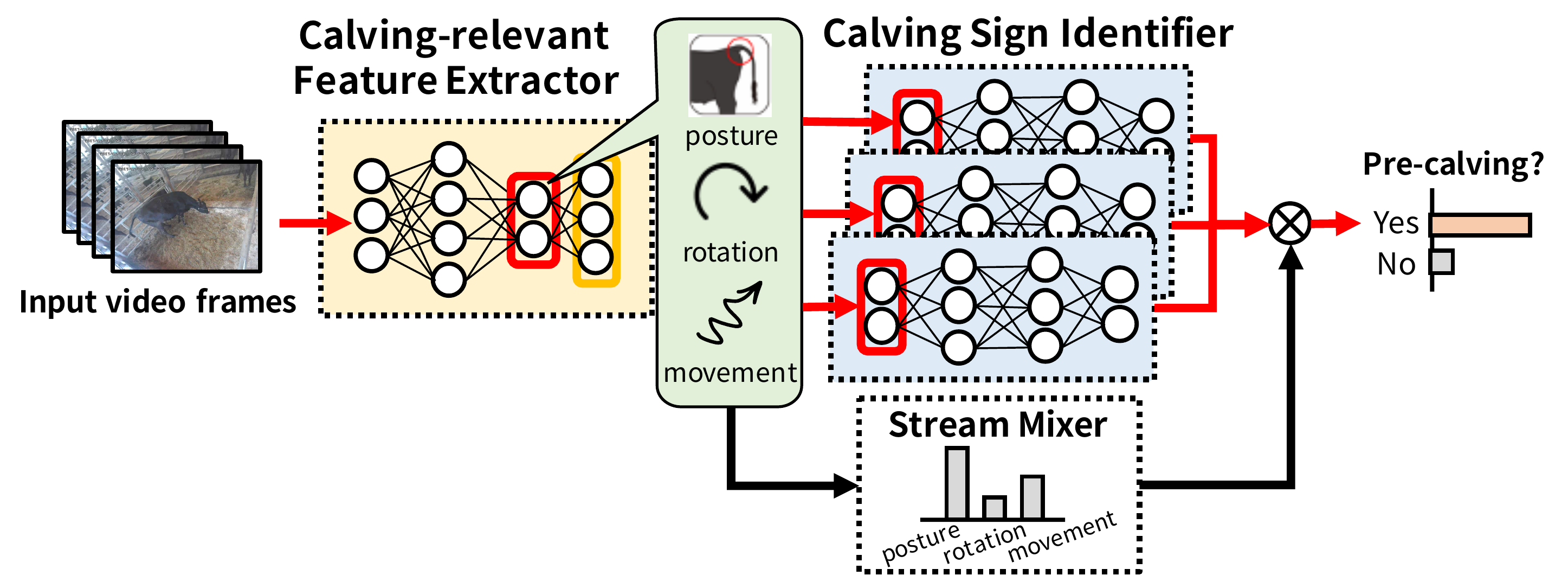}
\caption{Overview of proposed multi-stream network. Multiple calving-relevant features are extracted from video frames and used as inputs for each calving sign identifier. Each stream is weighted and integrated in accordance with situation.}
\label{fig:overview}
\end{figure}

When building a camera-based detection system, 
the \textit{end-to-end} approach with a single deep neural network (DNN) 
has become an indispensable technique
thanks to its simplicity when it comes to developing pattern recognition-based systems.
This approach is advantageous in that the feature extraction process can be learned in a single step.
In the case of real-world deployment
such as a calving detection system,
however, 
the \textit{end-to-end} approach is not appropriate because 
\textit{i)}
such an approach usually requires large-scale and organized data,
which is not always available in real-world deployment cases, 
and 
\textit{ii)}
the approach is \textit{black-box},
which is not feasible
to support farmers' decision-making
due to its lack of interpretability.

In the present study,
we aim to overcome the above challenges
of the \textit{end-to-end} approach
by incorporating the farmer's decision-making process into a neural network.
Farmers observe surveillance video
with a focus on attributes
such as pre-calving posture,
amount of rotation,
and amount of movement
to determine whether or not they are calving signs.
In this paper,
we propose a deep multi-stream neural network
in which such a decision-making process is embedded.
In our model,
individual streams that extract pre-calving information based on posture, rotation, and motion are constructed,
and their outputs are integrated 
in accordance with the situation
and then used to identify calving signs,
as shown in Fig.~\ref{fig:overview}.
The proposed system design is expected to solve the aforementioned problems 
in two ways:
\textit{i)} calving sign identification can be performed reliably
even with fewer calving scene data,
since only the calving-relevant information is extracted
from the first- and second-stage neural networks, and 
\textit{ii)} the basis for the predictor's decision can be provided to the farmers,
since individual streams and the stream mixture are designed 
on the basis of the experts' decision-making process.

To determine the effectiveness of the proposed multi-stream network,
we conducted experimental comparisons using video scenes
of 15 Japanese black beef cattle collected by a camera. 
Our findings should be useful 
in developing pattern recognition-based video surveillance systems 
that effectively incorporate experts' knowledge.

In Section \ref{sec:signs} of this paper, we briefly explain
the calving signs we used.
Section \ref{sec:model} describes the designed streams and their integration in the multi-stream architecture.
In Section \ref{sec:experiment}, we discuss the experiment and results.
We conclude in Section \ref{sec:conclusion} with a brief summary.

\section{Calving Signs Observable from Video}
\label{sec:signs}

This section presents an overview of the calving signs used.
Changes in the postures and actions of cows related to calving signs 
have been extensively investigated in animal science.

The typical posture-based calving signs that can be observed 
from images are as follows.
\begin{itemize}
\item\textbf{Switching between standing and lying postures}: About two to six hours before calving, the number of posture changes (\textit{e.g.,} switching between standing and lying) increases~\cite{Speroni2018, saint-dizier2015, Jensen2012} and the lying time increases two hours before calving~\cite{Jensen2012}. 

\item\textbf{Tail raising}: About four to six hours before calving, the frequency of tail raising increases~\cite{Jensen2012} and the position of the tail before calving is elevated~\cite{BUENO1981599, OWENS1985321}. 
\end{itemize} 
In this study, 
we examine the frequency of a cow standing, lying,
and raising her tail.


The action-based calving signs are observable
from video as follows.
\begin{itemize}
\item\textbf{Increase in the number of rotations and turns}: Characteristic walking patterns (\textit{e.g.,} rotations and turns) can be observed four hours before calving, and the observation frequencies increase, especially three hours before calving~\cite{Sugawara}. 

\item\textbf{Increase in aimless walking time}: The walking duration time on the actual calving day increases~\cite{saint-dizier2015, Jensen2012}. Aimless walking time apparently increases about 140 minutes before calving~\cite{OWENS1985321}. 
\end{itemize}
The proposed system exploits 
the statistics 
in the cows' rotation and movement behaviors.


\section{Calving Sign Detection Model}
\label{sec:model}

The proposed multi-stream network is composed
of three-stream processing 
for discriminative calving-relevant information extraction
and stream fusion
for calving sign identification.
The individual streams not only extract discriminative information
but also identify calving signs~(Anonymous 2020)
for each of the three types of calving-relevant features.
\begin{itemize}
    \item {\bf Posture-based feature}:
    The appearance of a cow standing, lying,
    and raising her tail are captured for each video frame using ResNet-50~\cite{7780459}
    and then accumulated into the relevant frequencies
    using temporal pooling techniques.
    
    \item {\bf Rotation-based feature}:
    Body direction information is extracted for each video frame using ResNet-50
    and accumulated into a statistic
    on the cow's rotation by measuring the changes in the body direction using the M-measure~\cite{Hermansky}.
        
    \item {\bf Movement-based feature}:
    The cow's region is detected for each video frame using YOLOv3~\cite{redmon},
    and differences in locations across frames are accumulated into a statistic on the cow's movement.
\end{itemize}

\begin{figure}[tb]
\centering
\includegraphics[width=\linewidth]{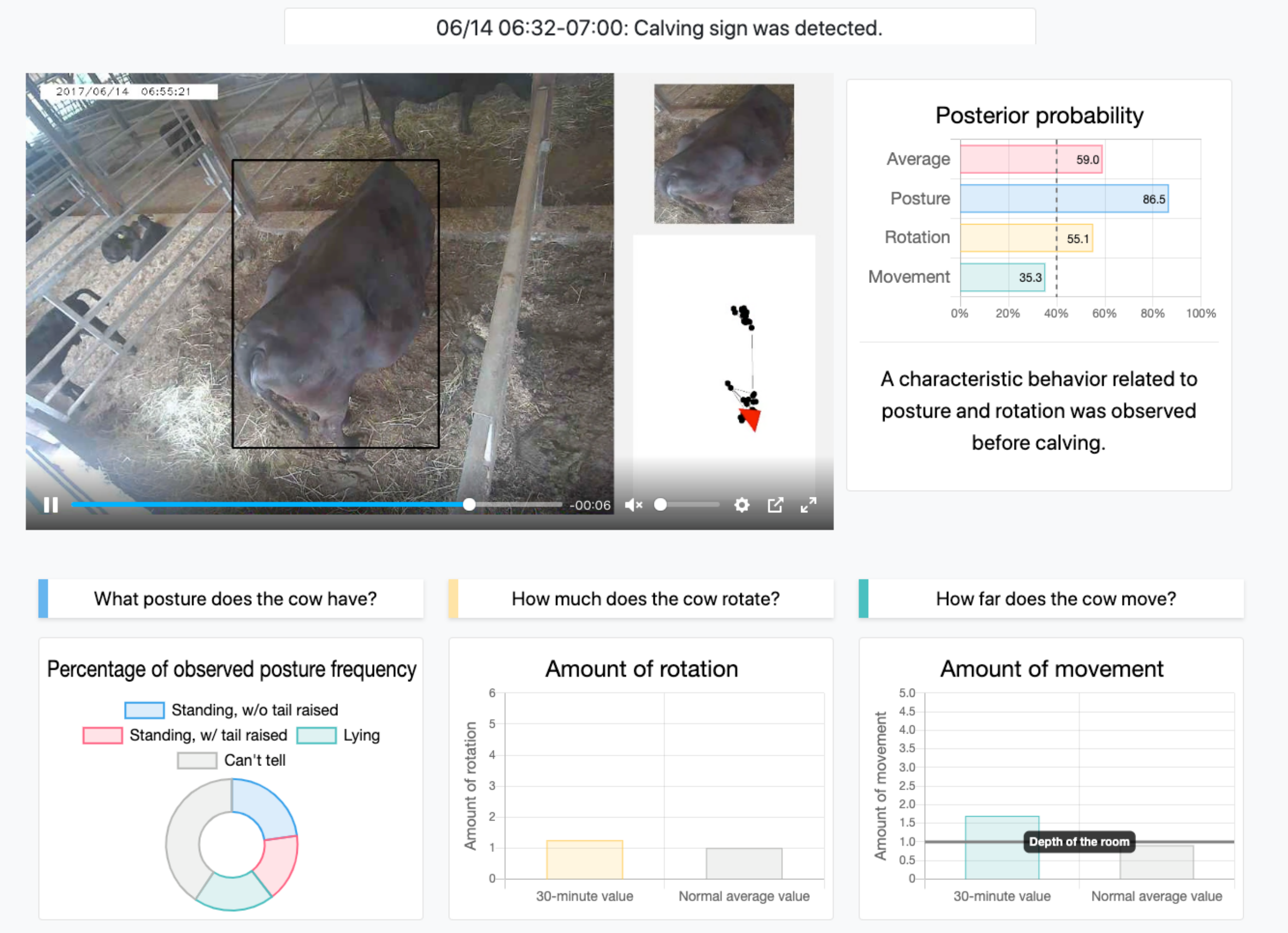}
\caption{User interface (UI) of proposed system. Top: predicted posterior probability. Bottom: frequency in pre-calving postures, statistics in rotation, and statistics in movement.}
\label{fig:ui}
\end{figure}
These features are extracted
using classifiers for a cow's status associated with calving signs. 
Note that
the classes to be identified are designed
such that they can be discriminated from video
without having to be an expert in animal husbandry.
In this case,
the annotation can be crowdsourced 
and thus the feature extraction process can be developed efficiently.
We should also emphasize that
the information obtained by the data-driven extraction of these calving-relevant features can be provided
as a basis for making predictive decisions.
Fig.~\ref{fig:ui} shows an example
of the user interface (UI)
of the developed system.
Not only the prediction results,
\textit{i.e.,} the predicted posterior probabilities
of calving signs, 
but also the statistics of a cow related to calving signs are presented
on the bottom of the screen.

The rest of this section describes the details
of calving-relevant feature extraction,
stream-level calving sign detection,
and multi-stream calving sign detection.
For feature extraction,
frame-by-frame features and time-series features are explained.

\subsection{Frame-wise Feature Extraction}

\begin{figure}[tb]
\begin{center}
\includegraphics[clip, width=\linewidth]{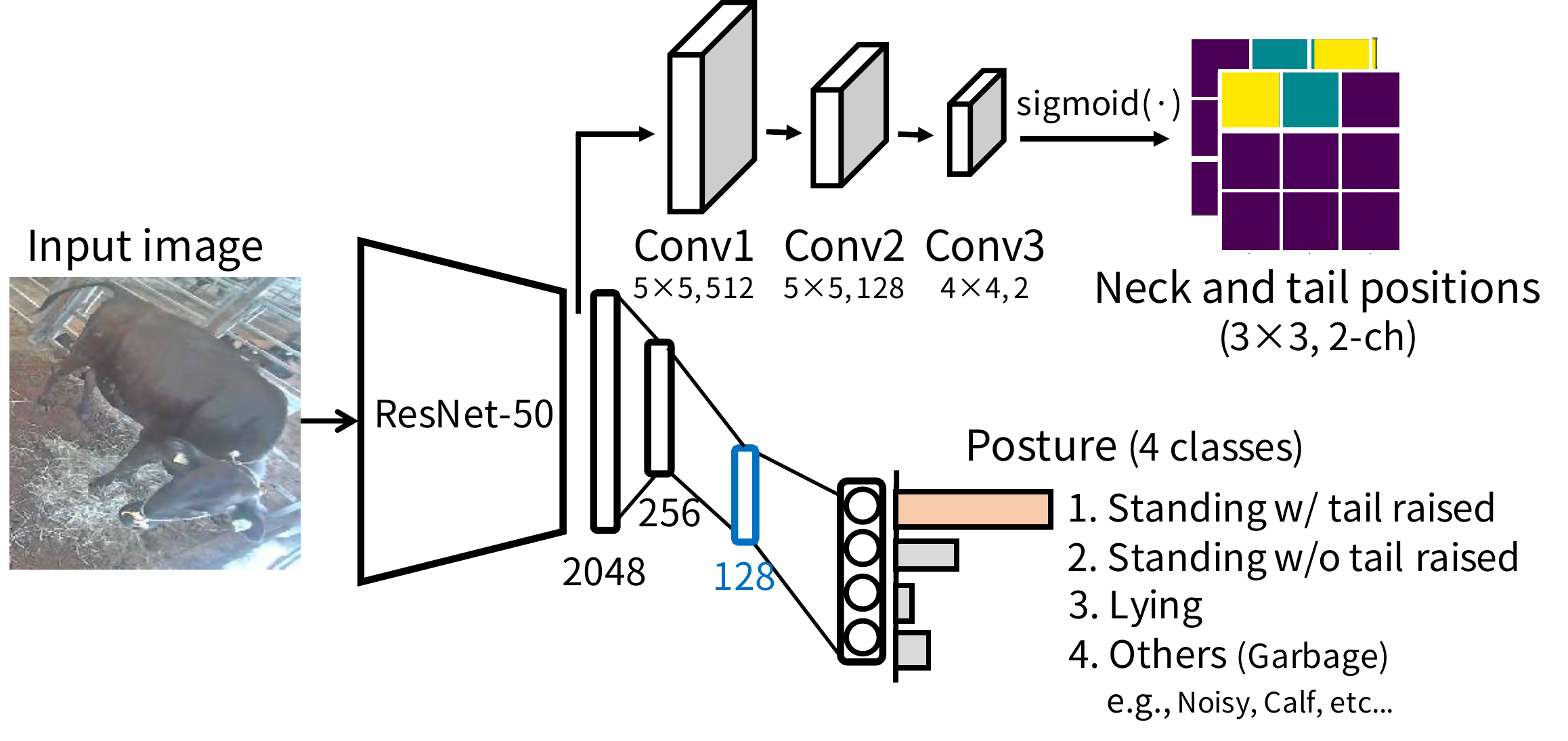}
\caption{Multi-task network architecture of frame-wise feature extractor. Hidden layer outputs (blue) convey posture information relevant to calving signs.}
\label{fig:extractor}
\end{center}
\end{figure}

The frame-wise information 
for the movement-based feature,
\textit{i.e.,} the cow's region,
is obtained using YOLOv3.
The frame-wise attributes
for the posture-based and rotation-based features are obtained
by a multi-task CNN
that identifies four classes of postures
and estimates the neck and tail positions.
The neck and tail positions implicitly convey information about a cow's body direction.
Fig.~\ref{fig:extractor} illustrates the architecture of this network.

Given an image, 
the posture classifier outputs the posterior probabilities 
of four classes:
1) standing cow with tail raised,
2) standing cow without tail raised,
3) lying cow,
and 4) others (garbage).
The garbage class includes images
that cannot be judged
(\textit{e.g.,} calf, no cows or noisy images).
The neck and tail position estimations yield
heatmaps for the root of the cow's neck
and of the cow's tail.
The convolutional network with the same structure as ResNet-50~\cite{7780459} is shared 
between both tasks.
The subsequent posture classification network 
is composed of 256 units of a fully connected layer and an output layer with four classes.
The neck and tail position estimation network is composed of multiple 2D convolutional layers,
and yields two-channel $3\times3$ posterior probabilities of the neck and tail existing
in the corresponding positions.
The objective function for training is $\mathcal{L} = \mathcal{L}_{\rm posture} + \mathcal{L}_{\rm position} + \mathcal{L}_{\rm ews}$,
where $\mathcal{L}_{\rm posture}$ and $\mathcal{L}_{\rm position}$ denote the cross-entropy loss for posture classification
and the binary cross-entropy loss
for position estimation,
respectively.
$\mathcal{L}_{\rm ews}$ is an element-wise sum of two estimated heatmaps,
which provides a constraint that the neck and tail cannot be in the same place.
Here, the $\mathcal{L}_{\rm ews}$ term yields an improvement in the accuracy and convergence speed.

As a preliminary experiment,
we examined the performance of this multi-task network
and found that 
the four postures were classified
with a 64.4\% accuracy,
while the neck and tail position were estimated
with 70.2\% and 67.6\% accuracies,
respectively.
While this network did not achieve a particularly high accuracy,
it is reasonable for use as a feature extractor.

\subsection{Time-series Feature Extraction}

\begin{figure*}[tb]
\begin{center}
\begin{tabular}{c}
\begin{minipage}{0.33\hsize}
\begin{center}
\includegraphics[clip, width=1.0\linewidth, height=\linewidth]{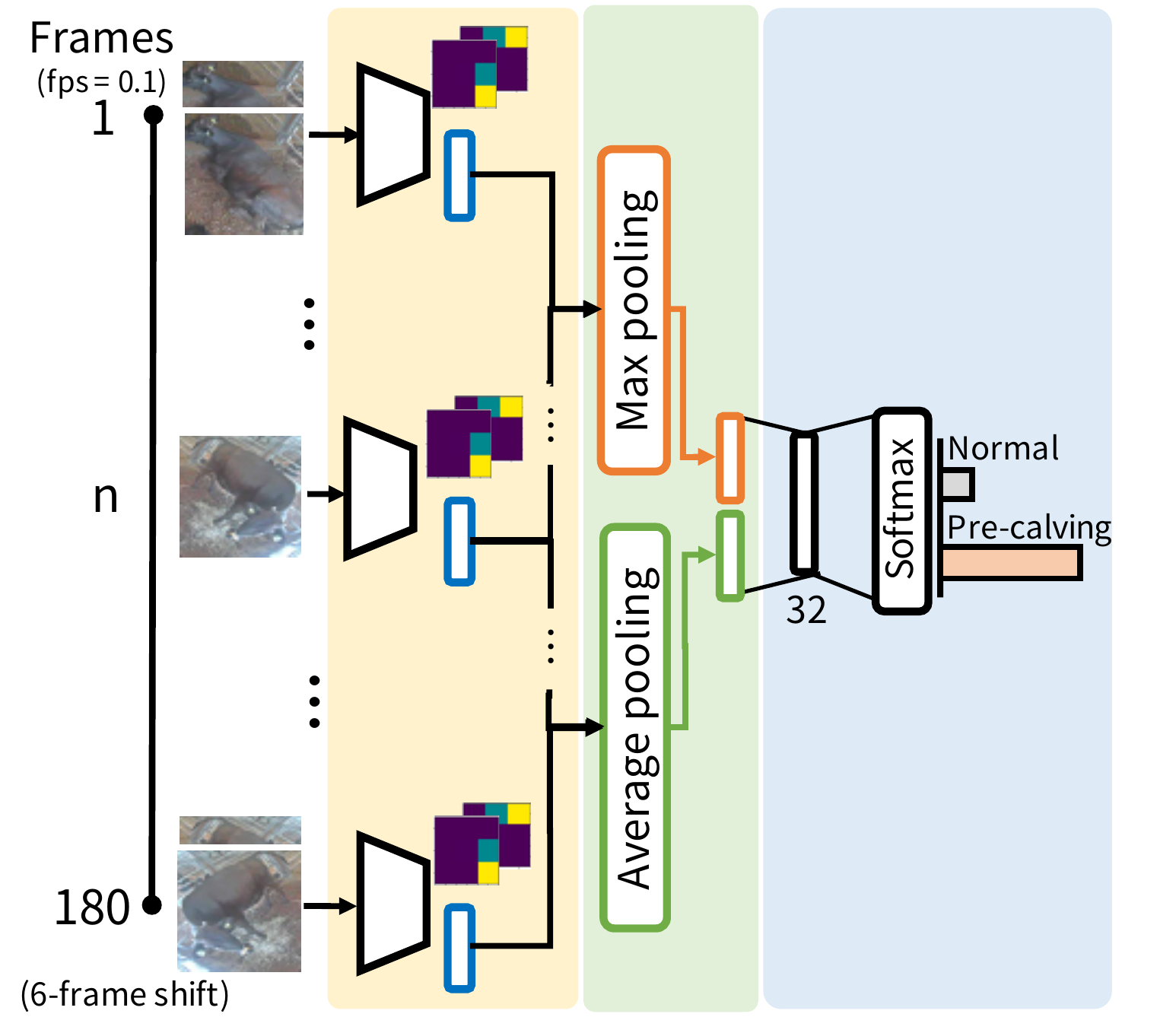}
\hspace{3cm} (a) Posture-based stream
\end{center}
\end{minipage}

\begin{minipage}{0.33\hsize}
\begin{center}
\includegraphics[clip, width=1.0\linewidth, height=\linewidth]{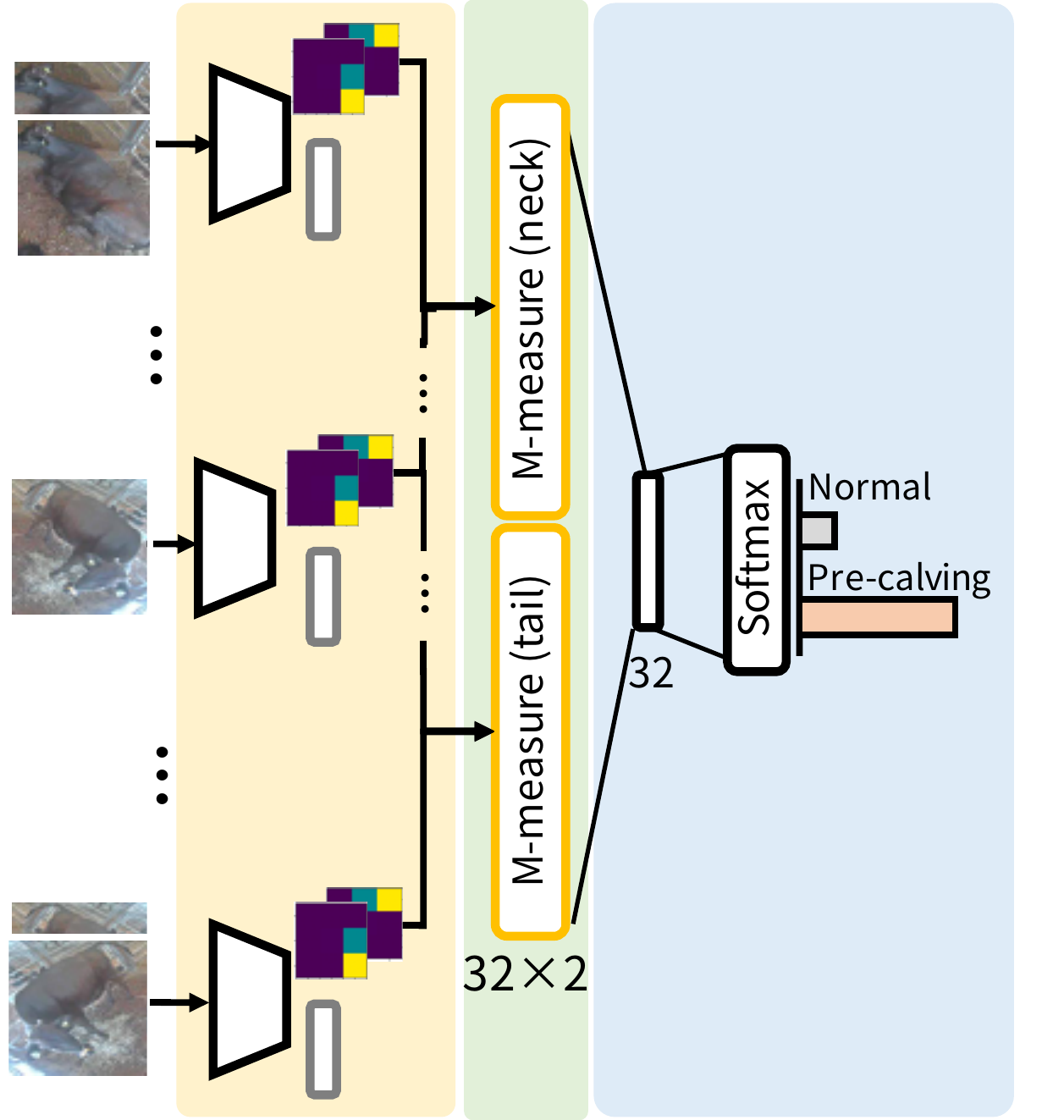}
\hspace{3cm} (b) Rotation-based stream
\end{center}
\end{minipage}

\begin{minipage}{0.33\hsize}
\begin{center}
\includegraphics[clip, width=1.0\linewidth, height=\linewidth]{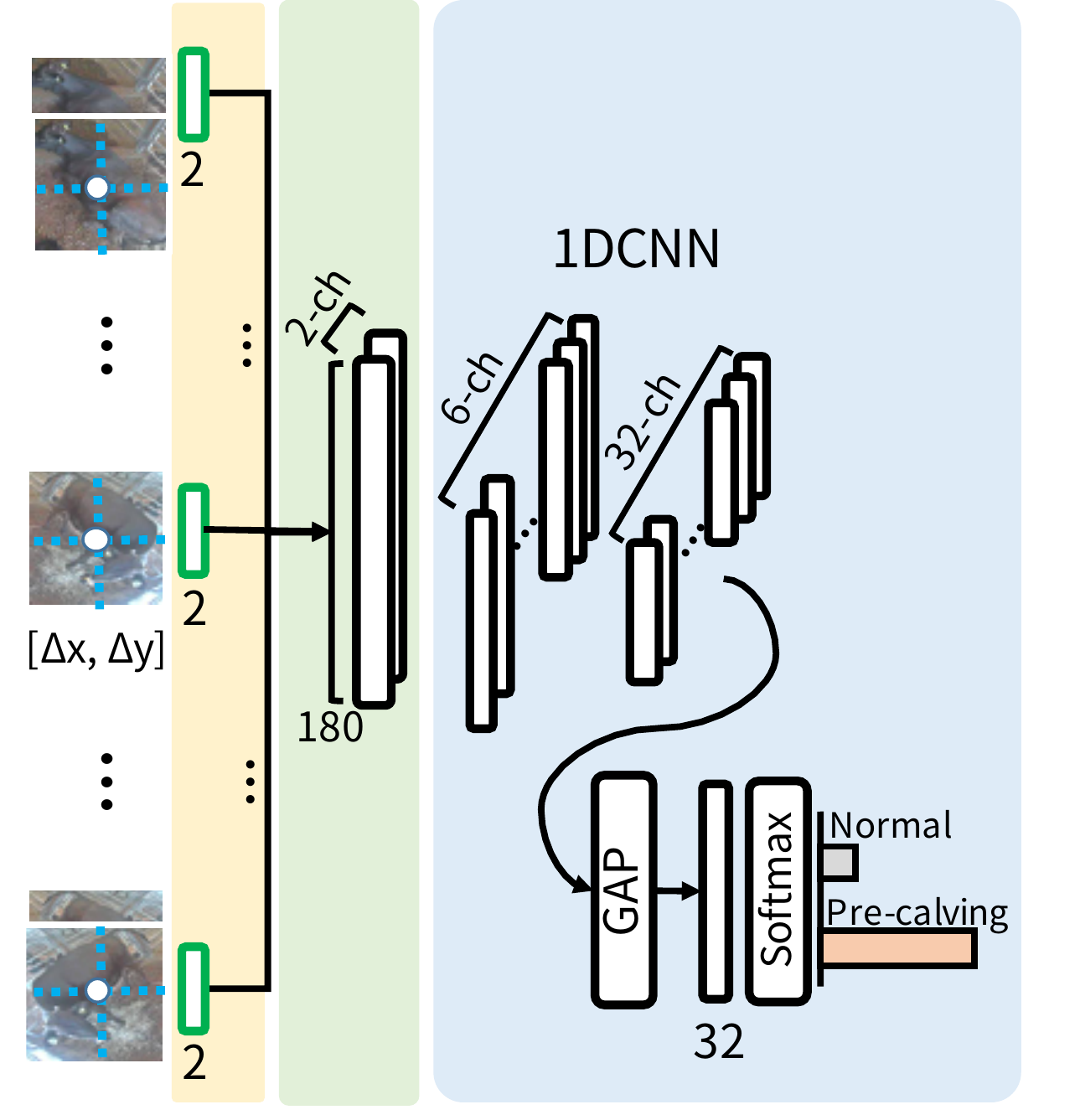}
\hspace{3cm} (c) Movement-based stream
\end{center}
\end{minipage}
\end{tabular}
\caption{Construction of streams composed of calving-relevant feature extractor and calving sign identifier. Yellow, green, and blue background processes represent frame-wise feature extraction, time-series feature extraction, and calving sign identification, respectively.}
\label{fig:detectors}
\end{center}
\end{figure*}

The above posture-based,
rotation-based,
and movement-based features are extracted
by accumulating the frame-wise features
described in the previous subsection.
Fig.~\ref{fig:detectors} schematically depicts 
the individual streams 
in which the corresponding calving-relevant features are extracted
and used to identify the calving signs.

Pre-processing was performed to compensate for frame dropping due to communication conditions.
We used mean-value interpolation and linear interpolation 
against masked features using the features of adjacent frames.
On the basis of the preliminary experimental results, 
we applied mean-value interpolation to the posture-based features 
and linear interpolation to the rotation-based and movement-based features.
These interpolations provided a significant contribution to the performance of the proposed systems.

\subsubsection{Posture-based stream}

This stream captures information
about the frequency in pre-calving postures.
The hidden layer outputs in the frame-wise posture classifier are accumulated
for a specified period (\textit{e.g.,} 180 frames)
into a single vector
using temporal max-pooling and average-pooling.
These vectors represent the frequency of the components that determine the posture of the cows. 
The output of max-pooling and that of average-pooling are concatenated
into the posture-based feature
and used as the input
for the posture-based calving sign identifier.

\subsubsection{Rotation-based stream}

This stream captures the frequency
in the changes of the cow's body direction,
which is determined by the neck and tail position of the cow.
Heatmaps (posterior probabilities) for the neck and tail positions are produced
by the frame-wise position estimator.
Here, 
the rotations and turns indicate significant changes in the body direction.
The M-measure~\cite{Hermansky, Ogawa} for the neck position and for the tail position can capture the frequency of the changes in the body direction. 
The M-measure is defined as the distance of the posterior probability ${\mathbf p}_{t \!-\! \Delta t}$
and ${\mathbf p}_{t}$
at a certain time interval ${\Delta t}$
as
\begin{equation}
\label{eq:def_M}
 \frac{1}{T-\Delta t} 
 \sum_{t=\Delta t}^{T}
 \mathcal{D}(p_{t-\Delta t}, p_{t}),
\end{equation}
where $\mathcal{D}(p,q)$ denotes the symmetric KL divergence
between the position posterior probability vectors,
which is defined as 
\begin{equation}
 \mathcal{D}({\mathbf p}, {\mathbf q}) = 
 \sum_{k=0}^K p^{(k)} \log \frac{p^{(k)}}{q^{(k)}} + 
 \sum_{k=0}^K q^{(k)} \log \frac{q^{(k)}}{p^{(k)}},
\end{equation}
where $p^{(k)}$ is the $k$-th element
of the posterior probability vector
$p\in \mathbb{R}^K$. 

\begin{figure}[tb]
\centering
\includegraphics[width=\linewidth]{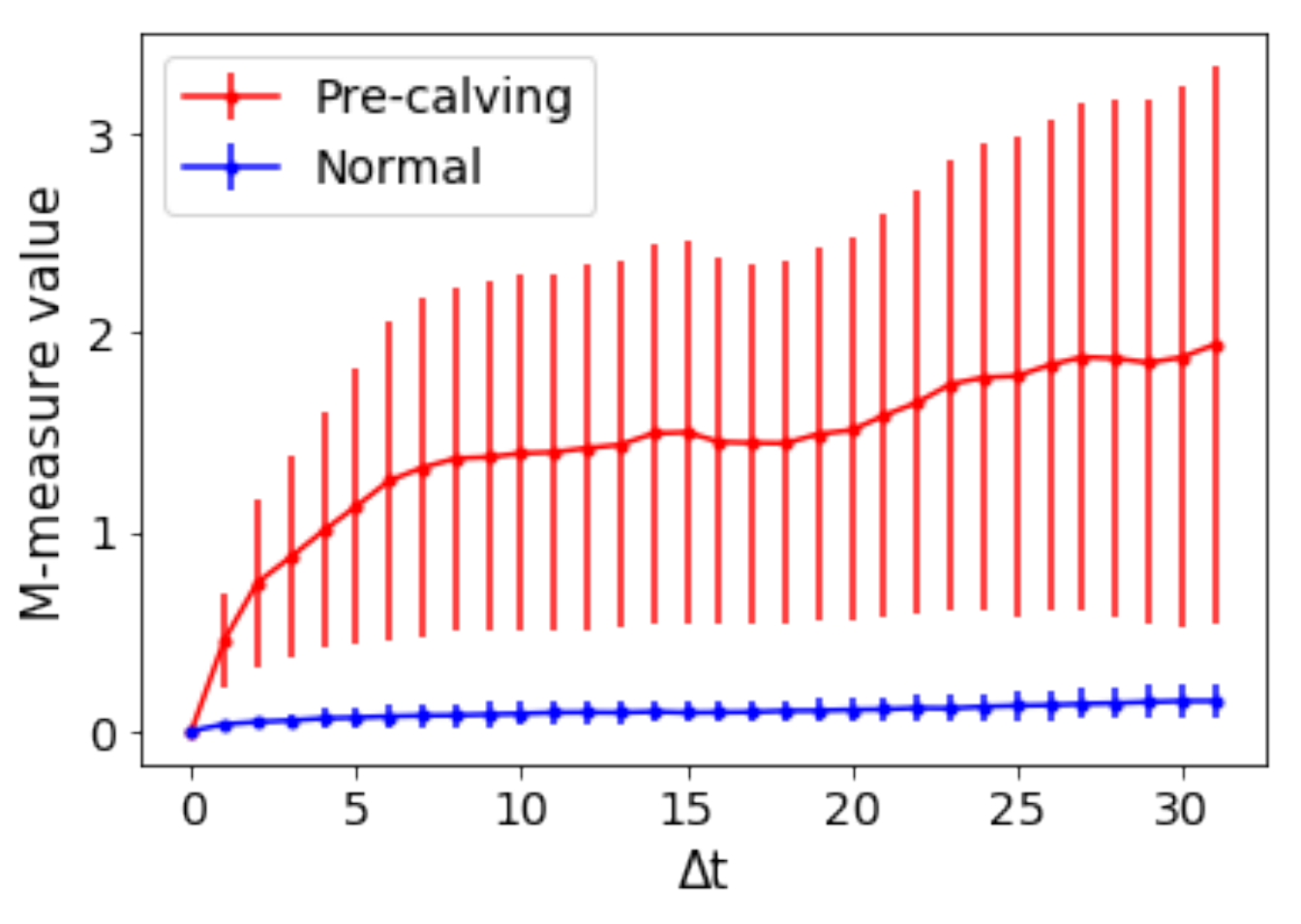}
\caption{M-measure values as function of $\Delta t$. 
        In this example, cow actually rotates before calving. 
        Video was recorded on May 27 and 28, 2017.}
\label{fig:mmeasure}
\end{figure}

Since there are fewer changes of a cow's body direction in the normal state,
the position posterior probability of the neck
and of the tail should not change.
Therefore, $\mathcal{M}(\Delta t)$ is low, regardless of $\Delta t$. 
In contrast,
since cows move around before calving,
the neck and tail positions change frequently with time.
Consequently, $\mathcal{M}(\Delta t)$ tends to be higher. 
From the above discussion,
we know that the M-measure can determine whether a cow is in a normal or pre-calving state. 
Fig.~\ref{fig:mmeasure} shows the average and standard deviation of the M-measure values
as functions of the time intervals $\Delta t$ ($\Delta t=1,...,32$)
for a particular cow.
We can see that
the M-measure pattern is distinctive
between the normal and pre-calving state,
when the cow actually rotates before calving.
In the present study,
the M-measure value is utilized
as a statistics in the cow's rotation, as
\begin{equation}
 \mathcal{M} = [\mathcal{M}(1), \mathcal{M}(2), \cdots, \mathcal{M}(32)].
\end{equation}
The M-measure value for the neck
and for the tail are concatenated
and used as the input
for the rotation-based calving sign identifier.

\subsubsection{Movement-based stream}

This stream accumulates the statistic
in the movement of a cow.
The cow's region (\textit{i.e.,} bounding box) is extracted
for each video frame using YOLOv3.
Here, an increase in the walking duration,
which is a known pre-calving characteristic, 
indicates a significant change in the cow's position.
The differences in the center coordinates 
of the cow's regions across video frames,
$\mathcal{C}_{t} = [x_{t} - x_{t-1}, y_{t} - y_{t-1}]$,
are concatenated 
for a specified period (\textit{e.g.,} 180 frames)
into the movement-based feature
and used as the input
for the movement-based calving sign identifier.

\subsection{Single-stream Calving Sign Detection}

Fig.~\ref{fig:detectors} illustrates the individual streams
based on different calving-relevant features.
For each stream,
time-series features are extracted
every 30 minutes (180 frames)
and taken as inputs
to the subsequent calving sign identifier.
The calving sign identifiers for the posture-based and rotation-based streams are built
using fully connected neural networks, 
which are composed of two fully connected layers with the ReLU~\cite{Nair} activation function.
The calving sign identifier for the movement-based stream is designed
using a 1D convolutional neural network 
that is composed of two 1D convolutional layers, 
a global average pooling (GAP) layer,
and two fully connected layers.

\subsection{Multi-stream Calving Sign Detection}

\begin{figure}[tb]
\centering
\includegraphics[width=\linewidth]{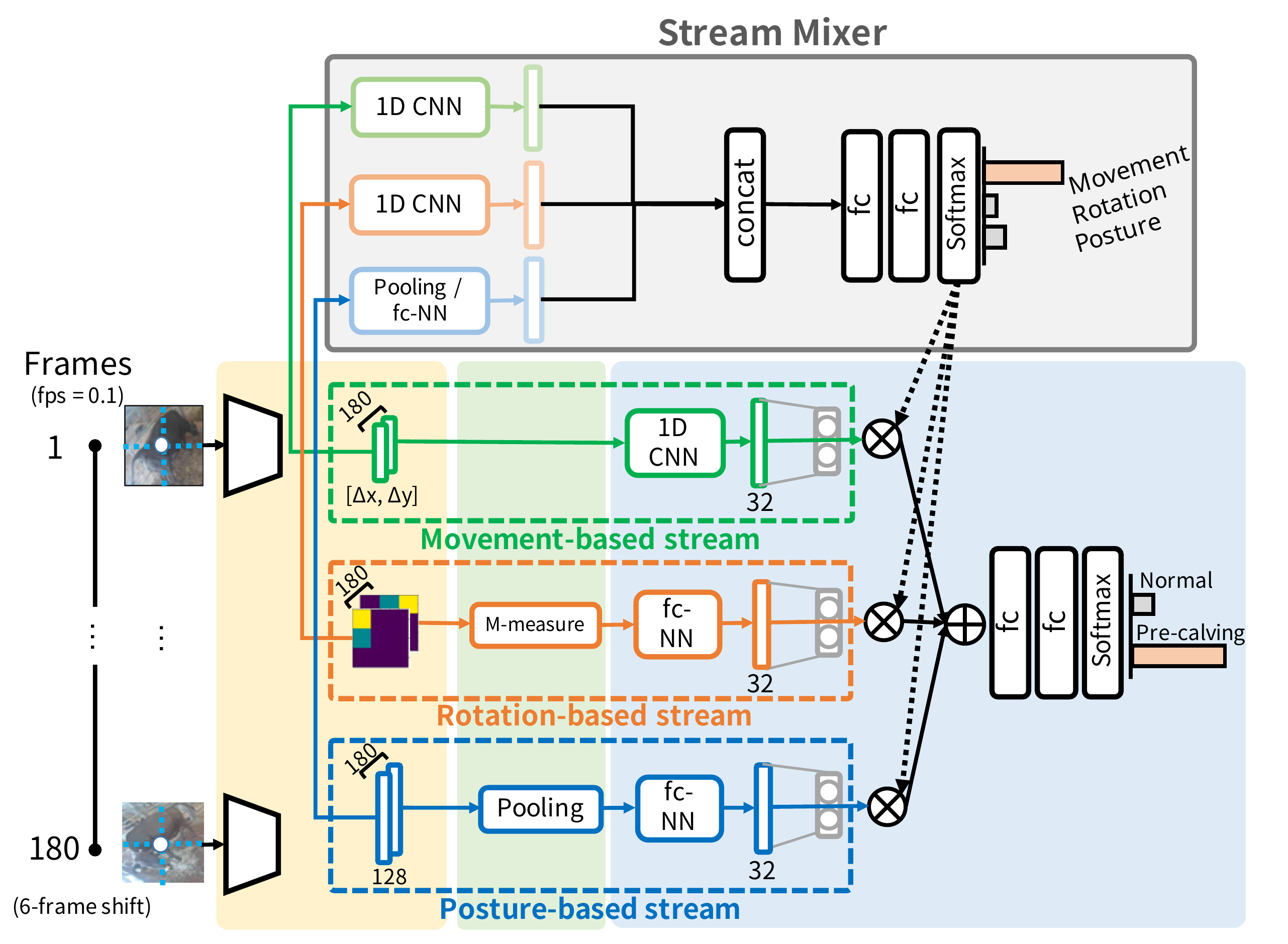}
\caption{Overview of proposed multi-stream network. Stream mixer adaptively
weighs and combines representations of single-stream systems. Mixture weights are calculated on basis of coordinate information, rotation-based features, and posture-based features.}
\label{fig:fusion}
\end{figure}

Individual streams based on different calving-relevant features are developed independently, as mentioned above.
Since each stream deals with different attribute features,
effective integration and selection of the developed streams should lead to further improvement.
Inspired by the architecture of mixture-of-experts~\cite{Shazeer},
we introduce a network for fusing the discriminative representations of each stream,
as shown in Fig.~\ref{fig:fusion}.
The stream mixer, 
which is composed of two fully connected layers with the ReLU activation function,
calculates mixture weights
for the individual streams (\textit{i.e.,} experts)
using the coordinate information,
rotation-based features,
and posture-based features.
The discriminative representations
for all streams
are then fused using the estimated mixture weights.
Here, the stream mixer decides the importance of attribute features
depending on the situation
in a data-driven manner.

The present study examines two types of information
as discriminative information
that contribute to the detection of calving signs:
the output of stream-level calving sign identifiers,
\textit{i.e.,} posterior probability of being pre-calving or normal,
and the hidden-layer outputs of the calving sign identifiers.
For the former,
the posterior probabilities for each stream are weighted
using the estimated stream weights
and then summed.
These weighted posterior probabilities are then used
for the final decision.
As for the latter,
32-dimensional outputs in the hidden layer
just before the output layer
can be considered
as a discriminative representation
to identify whether it is a pre-calving or normal.
These hidden representations for all streams are summed
with the estimated stream weights and 
then taken as an input to the final identifier
composed of two fully connected layers.

When training the multi-stream network, 
the weights of the three pre-trained streams are frozen 
and only the last feed-forward network and the stream mixer are updated
using the identification loss
of whether it is a pre-calving or normal state.

\section{Experiments}
\label{sec:experiment}

To determine the effectiveness of the proposed multi-stream calving detection system,
we conducted experimental comparisons using actual data from surveillance video.

\subsection{Data}

\begin{figure}[tb]
\centering
\includegraphics[width=0.9\linewidth]{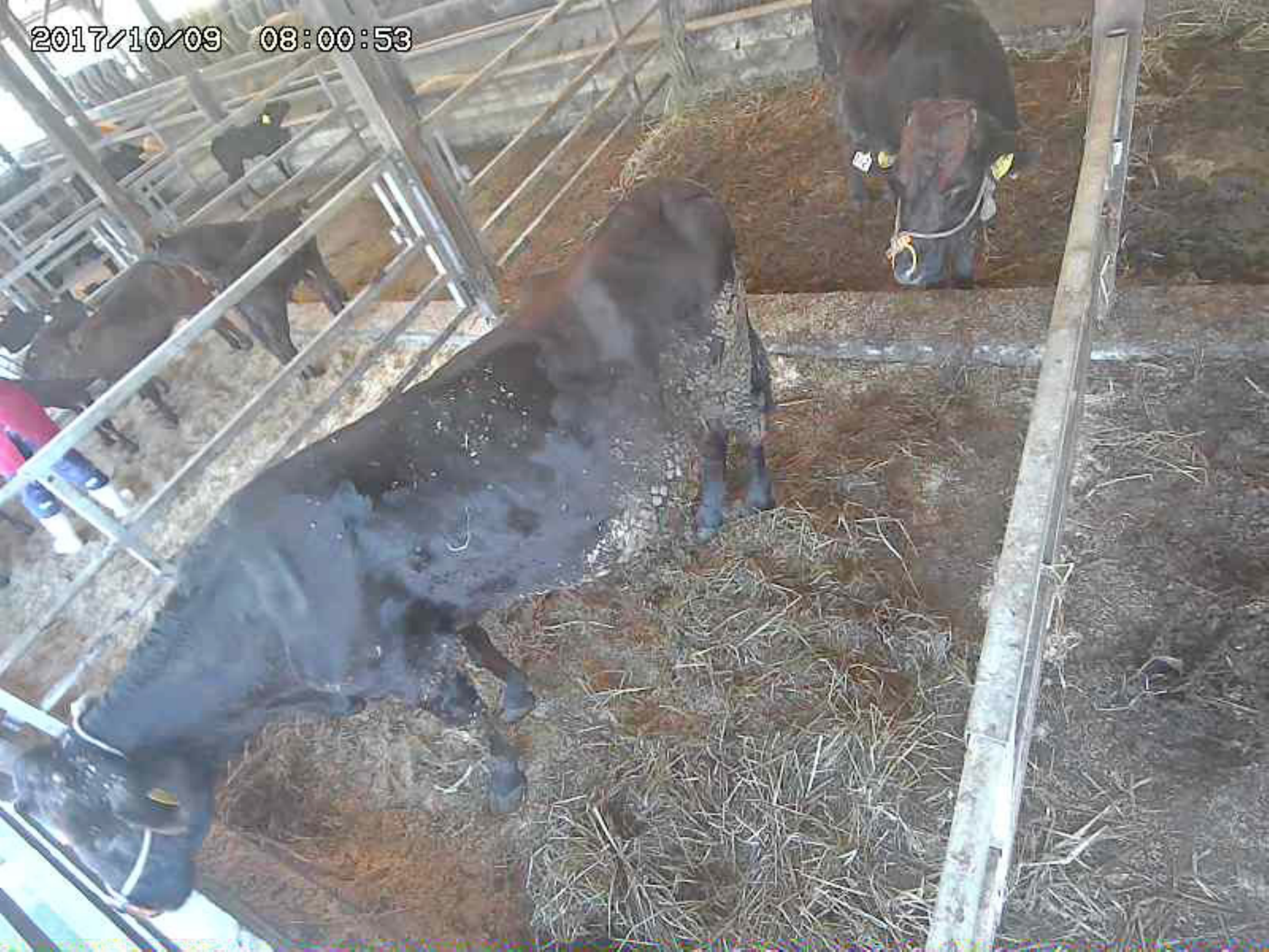}
\caption{\label{fig:cam-pic}Example of recorded frame.
Two pregnant cows are in same room.}
\label{fig:rec}
\end{figure}

The data were recorded on a farm 
in Kagoshima, Japan
from May to December 2017.
The recorded video included 15 different calving scenes
in which two pregnant Japanese black beef cattle are in a calving chamber.
Table~\ref{tab:data} lists the recording dates
of the experimental data.
Fig.~\ref{fig:rec} shows an example frame.
The scenes were acquired by a surveillance camera mounted in the chamber
and observed obliquely from above.
The frame rate was about 0.1.
To deal with dropped frames,
we applied the pre-processing described in Section \ref{sec:model}.
In this experiment,
the normal state was defined as 27 to 24 hours before calving,
and the pre-calving state was defined as three to zero hours before calving. 
The tracking of the targeted cow was conducted
on the basis of the intersection over union (IoU)
of the detected regions across adjacent frames.
We then fixed any erroneous tracking manually. 
\begin{table}[tb]
  \fontsize{9.0pt}{9.0pt} \selectfont
  \centering
  \begin{tabular}{c|c|c} \hline
    Index & Normal state & Pre-calving state \\ \hline \hline
    1 & 05/27 22:00 - 25:00 & 05/28 23:00 - 25:00 \\
    2 & 06/02 06:00 - 09:00 & 06/03 06:00 - 09:00 \\
    3 & 06/25 02:00 - 05:00 & 06/26 02:00 - 05:00 \\
    4 & 07/12 23:00 - 26:00 & 07/13 23:00 - 26:00 \\
    5 & 11/07 11:00 - 14:00 & 11/08 11:00 - 14:00 \\
    6 & 06/14 05:30 - 08:30 & 06/15 05:30 - 08:30 \\
    7 & 08/30 12:00 - 15:00 & 08/31 12:00 - 15:00 \\
    8 & 09/29 21:00 - 24:00 & 09/30 21:00 - 24:00 \\
    9 & 10/23 02:00 - 05:00 & 10/24 02:00 - 05:00 \\
    10 & 11/23 07:00 - 10:00 & 11/24 07:00 - 10:00 \\
    11 & 10/20 17:00 - 20:00 & 10/21 17:00 - 20:00 \\
    12 & 12/02 15:00 - 17:30 & 12/03 15:00 - 17:30 \\
    13 & 10/30 14:00 - 17:00 & 10/31 14:00 - 17:00 \\
    14 & 06/23 01:30 - 04:30 & 06/24 01:30 - 04:30 \\
    15 & 07/19 22:00 - 25:00 & 07/20 22:00 - 25:00 \\
    \hline
  \end{tabular}
  \fontsize{10.0pt}{10.0pt} \selectfont
  \caption{Dates of experimental data. All data were recorded in 2017.}
  \label{tab:data}
\end{table}

\subsection{Evaluation Items and Criteria}

First,
we compared the stream integration methods
in multi-stream calving sign detection.
Specifically, integration based on the final outputs 
(\textit{i.e.,} posterior probability) and intermediate layer outputs
of the stream-level calving sign identifier was examined.
The integration based on the posterior probability was performed as follows.
\begin{itemize}
\item\textbf{MS-Posterior-Average}:
The baseline system of posterior-based stream fusion,
which computes a simple average of the results 
of three individual streams.

\item\textbf{MS-Posterior-Mixer}:
Mixing the posterior probabilities
obtained from three individual streams
in accordance with the cow's situation
using a stream mixer.
The final output is the weighted sum of the weights
from the stream mixer~(Fig.~\ref{fig:fusion}) 
and posterior probabilities yielded from three individual streams.
\end{itemize}
The multi-stream integration based on the hidden layer outputs (HLOs) was performed as follows.
\begin{itemize}
\item\textbf{MS-HLO-Concat}:
The baseline system of HLO-based stream fusion
in which the hidden representations yielded from three streams are concatenated and taken as an input to a feed-forward neural network-based identifier.

\item\textbf{MS-HLO-Mixer}:
Mixing the hidden representations
yielded from three individual streams
according to the cow's situation
using a stream mixer,
as shown in Fig.~\ref{fig:fusion}.
\end{itemize}

Next,
the multi-stream system was compared
to the single-stream
and the \textit{end-to-end} (E2E) systems
as follows.
\begin{itemize}

\item\textbf{SS-Posture}:
Single-stream calving detection system
with posture-based feature extraction.

\item\textbf{SS-Movement}:
Single-stream calving detection system
with movement-based feature extraction.

\item\textbf{SS-Rotation}:
Single-stream calving detection system
with rotation-based feature extraction.

\item\textbf{E2E}:
A \textit{black-box} deep neural network
without explicit calving-relevant feature extraction.

\end{itemize}
The {\bf E2E} system has the same architecture 
as the posture-based stream
shown in Fig.~\ref{fig:detectors}~(a). 
Frame-level features were extracted
using Resnet-50 pre-trained on ImageNet~\cite{ImageNet}
and then pooled as time-series patterns.

The above systems were evaluated
on the basis of four criteria:
precision,
recall,
F1-measure,
and area under the curve~(AUC).
The precision and recall rates were calculated
with pre-calving as a positive case.
A 15-fold nested cross-validation test was carried out
for the evaluation.

\subsection{Experimental Setup}

All experiments were implemented in Pytorch~\cite{pytorch}.
The stochastic gradient descent was used to train all models, 
and the learning rate was set to 0.005.
The objective function was the cross-entropy loss.
The learning rate was reduced
by a factor of five
when an increase in the validation loss was observed
in two consecutive runs.
Early stopping was applied
when two consecutive increases in the validation loss were observed shortly
after the learning rate was updated. 
The mini-batch size was 20. 

\subsection{Experimental Results}

\begin{figure}[tb]
\centering
\includegraphics[width=\linewidth]{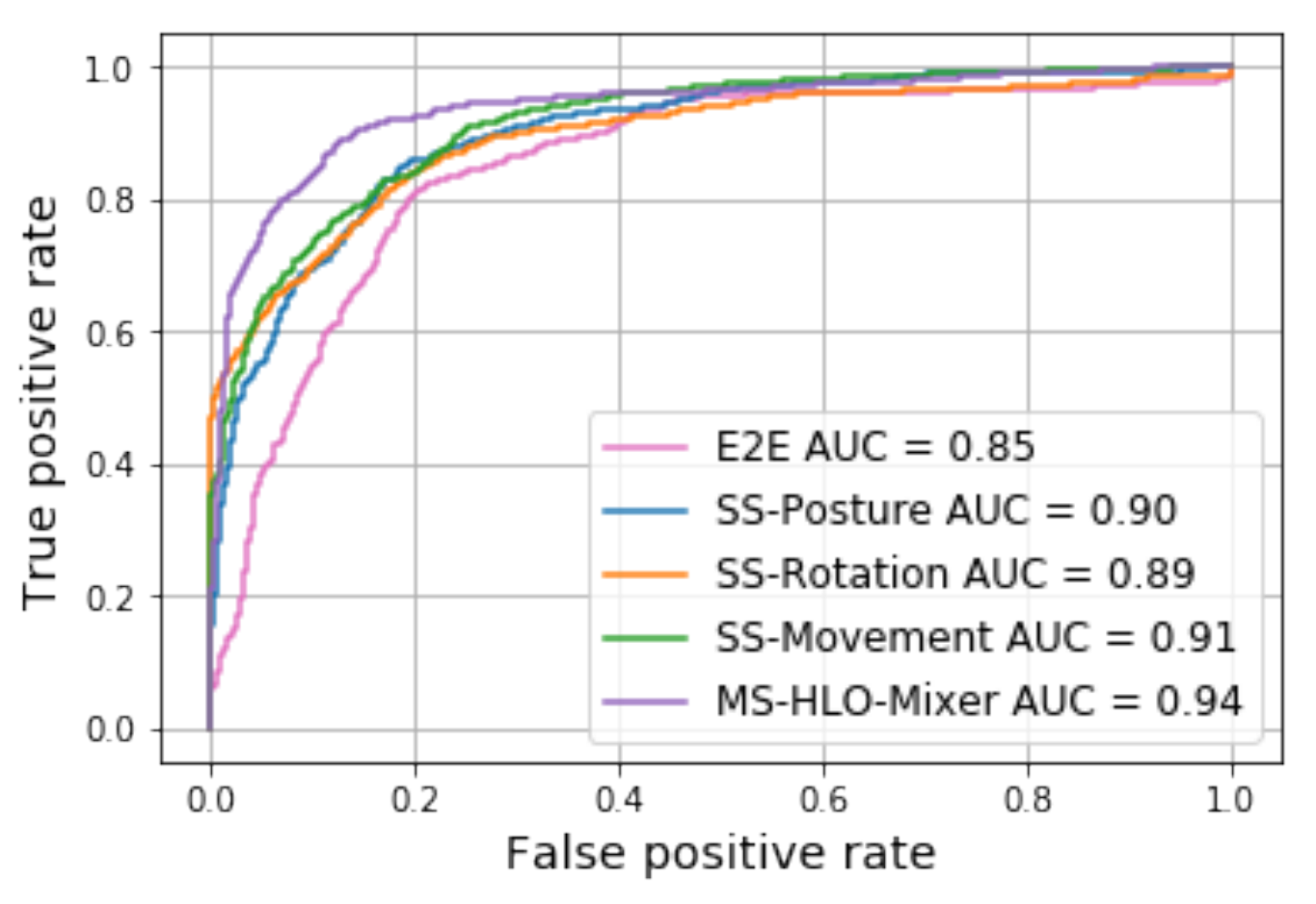}
\caption{ROC curves for end-to-end system, single-stream systems, and best multi-stream system (MS-HLO-Mixer).}
\label{fig:roc}
\end{figure}

Table~\ref{tab:fusion_comparison} lists the precision, recall, F1-measure, and AUC values
for the different methods of fusing multiple streams
on all testing data. 
Table~\ref{tab:predict-result} lists the precision, recall, F1-measure, and AUC values
for the proposed multi-stream system,
single-stream systems, 
E2E system,
and other stream selection techniques
(such as the upper limit,
max probability selection,
and minimum probability selection)
on all testing data. 
Fig.~\ref{fig:roc} shows the ROC curves.
Overall, the results indicate that
the single and multi-stream systems with explicit calving-relevant feature extraction yielded improvements over the \textit{black-box} E2E system.
In terms of average performance, 
the \textbf{MS-HLO-Mixer} system performed the best.
\begin{table}[tb]
\centering
 \vspace{0.5cm}
 \begin{tabular}{l|cccc} \hline
     Stream fusion & Precision & Recall & F1 & AUC \\
     \hline \hline
     MS-Posterior-Average & 0.83 & 0.88 & 0.86 & 0.93 \\
     MS-Posterior-Mixer & 0.83 & 0.87 & 0.85 & 0.91 \\
      \hline
     MS-HLO-Concat & 0.85 & 0.87 & 0.86 & 0.93\\
     MS-HLO-Mixer & \bf{0.86} & \bf{0.89} & \bf{0.88} & \bf{0.94}\\
   \hline
  \end{tabular}
  \fontsize{10.0pt}{10.0pt} \selectfont
  \caption{Precision, recall, F1-measure, and AUC values yielded by different methods of fusing multiple streams on all testing data. }
  \label{tab:fusion_comparison}
\end{table} 
\begin{table}[tb]
\centering
 \vspace{0.5cm}
 \begin{tabular}{l|cccc} \hline
      & Precision & Recall & F1 & AUC \\ \hline \hline
     Upper Limit & 0.91 & 0.93 & 0.92 & 0.98 \\ 
     Max Prob Selection & 0.73 & 0.93 & 0.82 & 0.92\\
     Min Prob Selection & 0.89 & 0.73 & 0.80 & 0.91\\\hline
     E2E & 0.73 & 0.86 & 0.79 & 0.85\\\hline
     SS-Posture & 0.81 & 0.85 & 0.83 & 0.90 \\
     SS-Rotation & 0.80 & 0.84 & 0.82 & 0.89 \\
     SS-Movement & 0.81 & 0.83 & 0.82 & 0.91 \\\hline
     MS-HLO-Mixer & \bf{0.86} & \bf{0.89} & \bf{0.88} & \bf{0.94}\\
   \hline
  \end{tabular}
  \fontsize{10.0pt}{10.0pt} \selectfont
  \caption{Precision, recall, F1-measure, and AUC values yielded by end-to-end system, three single-stream systems, best multi-stream system (MS-HLO-Mixer), and other reference values
 on all testing data.}
  \label{tab:predict-result}
\end{table} 
\begin{figure*}[tb]
\centering
\includegraphics[width=\linewidth]{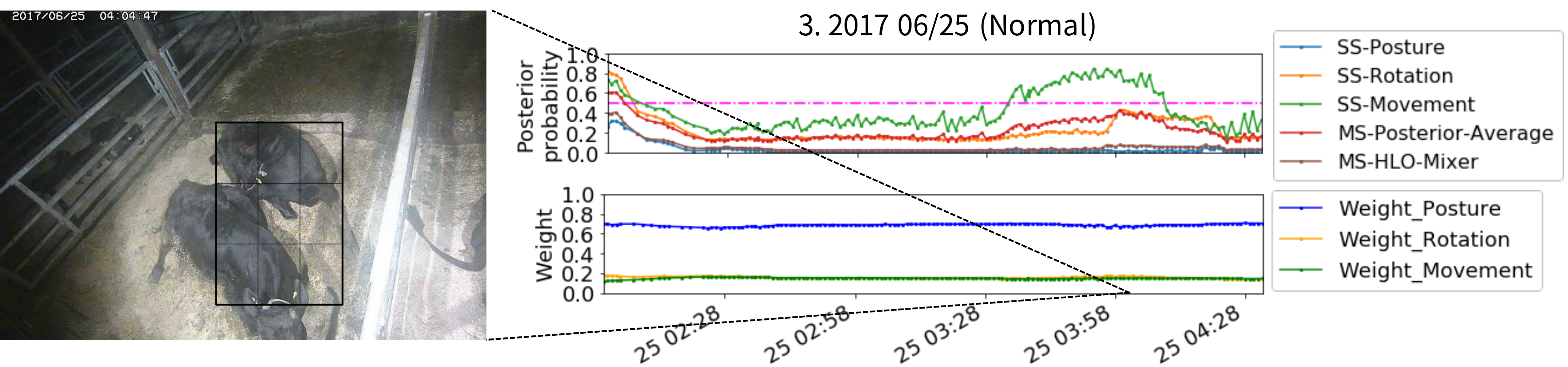}
\caption{Transition of posterior probabilities and mixture weights of MS-HLO-Mixer system in scene of normal state. Simple averaging (MS-Posterior-Average) was affected by SS-Movement stream that incorrectly perceived frequent changes in cow’s position (left). In contrast, MS-HLO-Mixer system gave large weight on SS-Posture stream, which suppressed false positives.}
\label{fig:pb_trans}
\end{figure*}
\subsubsection{Comparison of stream fusion}

Table~\ref{tab:fusion_comparison} shows that
the \textbf{MS-HLO-Mixer} system achieved a superior performance
compared to the other multi-stream systems.
The posterior probability-based fusion possibly suffered poor representation
due to its low dimension.
For HLO-based fusion,
\textbf{MS-HLO-Concat},
which is a commonly applied technique thanks to its simplicity,
yielded a strong bias towards the posture stream,
while the \textbf{MS-HLO-Mixer} system did not.
The stream mixer contributed
to reducing the information redundancy
that makes a particular stream dominant.


\subsubsection{Comparison of na\"ive stream selection}
\label{subsec:comp_selection}

The Max Prob Selection and Min Prob Selection in Table \ref{tab:predict-result}
selected the streams that yielded the largest and smallest probability,
respectively,
out of the individual streams.
In other words, these stream selection methods
could reduce false positives or false negatives
in a simple way.
There was a large bias towards precision or recall
resulting in a significant drop in F1-measure.
In contrast,
the \textbf{MS-HLO-Mixer} system improved the performance
while maintaining a balance between the precision and recall.

\subsubsection{Complementarity of individual streams}
\label{subsec:comp}

Table \ref{tab:predict-result} shows the calculated upper limit of the detection performance.
This value was calculated 
by choosing the best stream
when the true state was known~(\textit{i.e.,} cheating):
the stream yielding the highest posterior probability was chosen
if the true state was pre-calving,
while the stream yielding the lowest posterior probability was chosen
if the true state was normal.
As we can see,
the upper limit performed significantly better in terms of all metrics, 
indicating that the 
three streams give complementary predictions.
The proposed \textbf{MS-HLO-Mixer} system
yielded the high performance 
that was closest to the upper limit.

\subsubsection{Analysis of stream weights}
\label{subsec:analysis_w}

Visualizing the mixture weight patterns
in the \textbf{MS-HLO-Mixer} system enables us to confirm
which attributes the system considers important,
depending on the situation.
Fig.~\ref{fig:pb_trans} shows the transition
of the posterior probabilities and mixture weights
for one sequence of a normal state.
This was a late-night scene
where two cows were continuously lying.
Since the two cows were lying close to each other,
the detected region of the targeted cow changed significantly across frames.
In this case,
the \textbf{SS-Movement} system incorrectly perceived
frequent changes in the cow's position,
which yielded unduly high probability.
As a result,
the \textbf{MS-Posterior-Average} system,
which simply averaged the posterior probabilities
of individual streams,
tended to unduly output high posterior probability.
In contrast,
the proposed \textbf{MS-HLO-Mixer} was heavily weighted
to the \textbf{SS-Posture} system,
as shown in ``Weight\_Posture''
of Fig.~\ref{fig:pb_trans},
and correctly suppressed false positives,
the same as the \textbf{SS-Posture} system.
These results demonstrate that 
the mixture weights can be used
as an indicator of which attributes to focus on,
which will aid farmers with their decision-making.

\section{Conclusion}
\label{sec:conclusion}

We have developed a deep multi-stream network that effectively detects calving signs in cows.
Farmers typically determine whether cows are near to calving on the basis of the characteristics of different attributes such as pre-calving posture, amount of rotation, and movement.
The proposed system design incorporates such a decision-making process 
and works well to overcome the drawbacks of the \textit{end-to-end} approach without explicit feature extraction.
Our results demonstrate that the proposed multi-stream network
can reduce detection errors compared to the baseline systems.
Moreover, the mixture weights can be used as 
an indicator to understand the behavior of the system, 
thus aiding farmers with decision-making.
In the broader scope, our findings contribute insights into how to build
pattern recognition-based video surveillance systems that 
effectively incorporate experts' knowledge.


\newpage

\begin{quote}
\begin{small}
\bibliography{library}
\end{small}
\end{quote}

\bigskip
\end{document}